%% file: acl_latex.tex
\pdfoutput=1

\documentclass[11pt]{article}

\usepackage{acl}

\usepackage{multirow}
\usepackage{graphicx}
\usepackage{pifont}
\usepackage{amssymb}
\usepackage{caption, subcaption}
\usepackage{array}
\usepackage{booktabs}
\usepackage{times}
\usepackage{latexsym}
\usepackage{comment}
\usepackage[T1]{fontenc}

\usepackage[utf8]{inputenc}

\usepackage{microtype}

\title{Conversation Derailment Forecasting with Graph Convolutional Networks}

\author{Enas Altarawneh \\   York University \\
\texttt{enas@eecs.yorku.ca}  \\ \And 
  Ameeta Agrawal \\
  Portland State University\\
\texttt{ameeta@pdx.edu} \\ \vspace{-15pt}
  \\\AND
  Michael Jenkin \\
  York University\\
\texttt{jenkin@eecs.yorku.ca} \\
  \\\And
  Manos Papagelis \\ 
  York University\\ \texttt{papaggel@eecs.yorku.ca} \\
  }

\begin{document}
\maketitle
\begin{abstract}
{Online conversations are particularly susceptible to derailment, which can manifest itself in the form of toxic communication patterns like disrespectful comments or verbal abuse. Forecasting conversation derailment  predicts signs of derailment in advance enabling proactive moderation of conversations. Current state-of-the-art approaches to address this problem rely on sequence models that treat dialogues as text streams. We propose a novel model based on a graph convolutional neural network that considers dialogue user dynamics and the influence of public perception on conversation utterances. Through empirical evaluation, we show that our model effectively captures conversation dynamics and outperforms the state-of-the-art models on the CGA and CMV benchmark datasets by 1.5\% and 1.7\%, respectively.}

\end{abstract}

\input{introduction.tex}

\section{Related Work}
 \label{sec:related}
There has been considerable research attention on the problem of detecting various forms of toxicity in text data. There are methods for identifying cyberbullying \cite{abuse}, hate speech \cite{hate}, or negative sentiment \cite{agrawal2014kea,sent} or lowering the intensity of emotions \cite{ziems2022inducing,xie23}. These methods are useful in filtering unacceptable content. However, the focus of these models is on mostly analyzing or classifying already posted harmful texts.


The CRAFT models introduced by \citet{chang2022thread} are the first models to go beyond classification of hate speech to addressing the problem of forecasting conversation derailment.  The CRAFT models integrate two components: (a) a generative dialog model that learns to represent conversational dynamics in an unsupervised fashion, and (b) a supervised component that fine-tunes this representation to forecast future events. As a proof of concept, a mixed methods approach combining surveys with randomized controlled experiments investigated how conversation forecasting using the CRAFT model can help users \cite{chang2022thread} and moderators \cite{schluger2022proactive} to proactively assess and deescalate tension in online discussions. 

Extending the hierarchical recurrent neural network architecture with three task-specific loss functions proposed by \citet{time} was shown to improve the CRAFT models. After pretrained transformer language encoders such as BERT \cite{bert} proved to be successful at various NLP tasks, \citet{dynamic} explored how they can be used for forecasting derailment. The  model in this work  consists of  a BERT checkpoint with a sequence classification (SC) head. Similarly, 
\citet{de2021beg} evaluated feature-based and neural models to predict whether disagreements in Wikipedia Talk page conversations will be escalated to mediation by a moderator. 


\citet{saveski2021structure} studied the relationship between structure and toxicity in conversations on Twitter at individual, dyad, and group level, and found that social relationships among users influence their behaviors. \citet{salehabadi2022user} also studied the differences between toxic and non-toxic conversations on Twitter, highlighting important differences between user engagement and toxicity. While these recent works stress the importance of user characteristics in conversation modeling, to our knowledge, models that incorporate such signals for the task of predicting derailment have remained unexplored. 

Here we propose graph-based models for leveraging user-specific information. Graph convolutional neural networks  have been used for conversation classification. In one popular application,  emotion estimation, the graph model is used to account for speaker related information \cite{DGCN,DisGNN}.  Other  work have used similar graph neural networks to  forecast  emotions   \cite{affect-rich,PositiveEE,HCNN, predict}.

\section{Problem Definition}
 \label{sec:problem}
In this section, we formally define the problem of {\em forecasting conversation derailment}. For a conversation  $\mathcal{C}=\{\{t_1,t_2,...,t_N\}\mathbin{,}\{u_1,u_2,...,u_N\}\mathbin{,}\{s_1,s_2,...,s_N\}\}$ consisting of $N$ turns, the last turn (i.e., the $N$'th turn) is the potential site of derailment where $l=\{civil,personal$ $ attack\}$ denotes the label of this turn.  

For the $i$th turn, $t_i$ denotes its text, $u_i$ denotes its user, and $s_i$ denotes an optional score, e.g., number of votes (up-vote/down-vote). An up-vote is a positive impression and a down-vote is a negative impression on the turn utterance. The goal is to forecast the derailment label $l$ of the $N$'th turn given a conversation  $\mathcal{C}$ up to $N-1$ turns (i.e., without any information about the $N$th turn).


\section{Model for Forecasting Conversation Derailment}

\label{sec:models}
\begin{figure*}[h]
   \centering\includegraphics[width= 16cm]{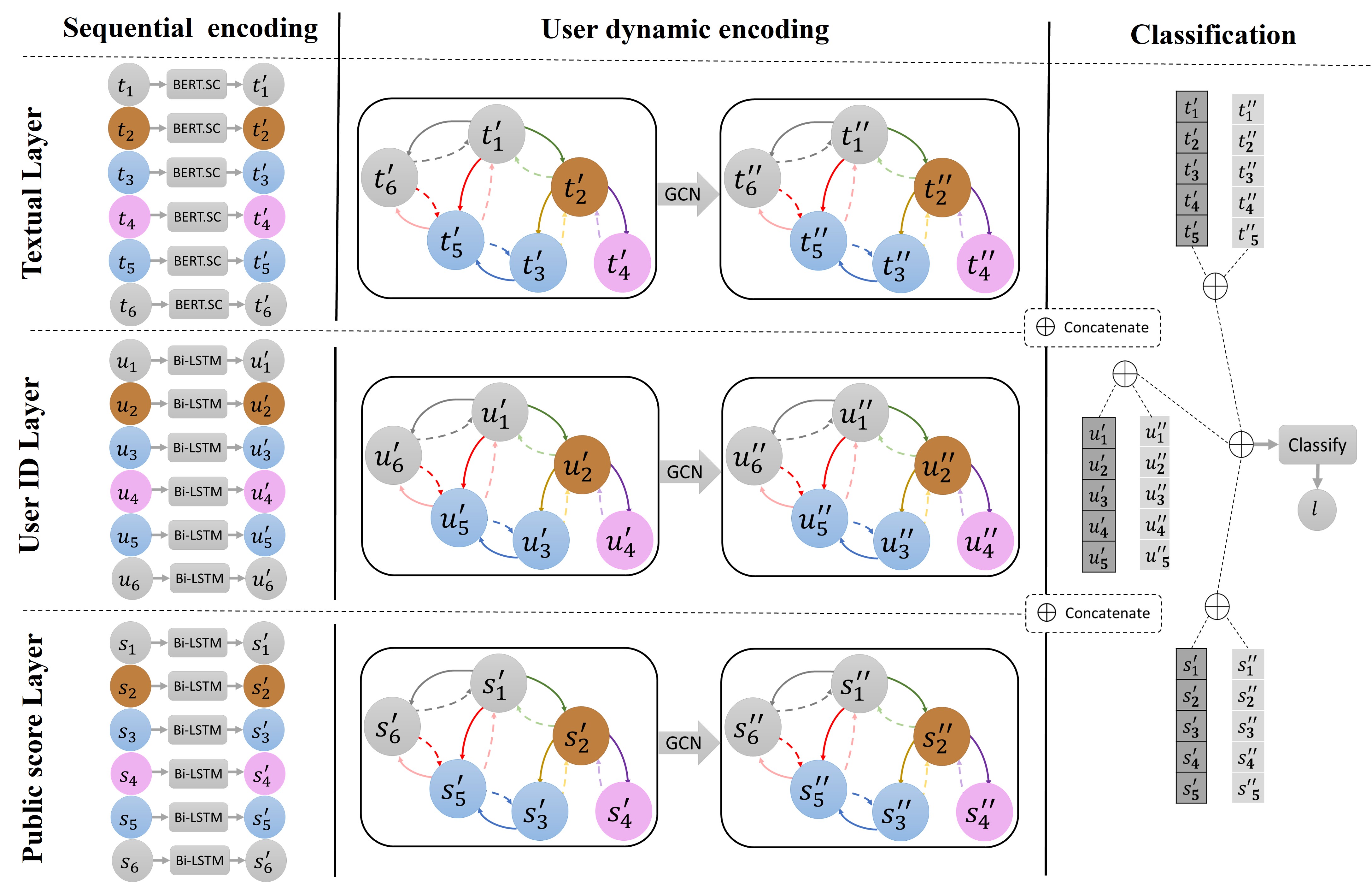}
   \caption{The  FGCN model architecture.}
   \label{GCN}
  
\end{figure*}

In this section, we describe our proposed Forecasting Graph Convolutional neural Network model, (FGCN),  visualized in Figure~\ref{GCN}. 

\subsection{Sequential encoding}
The input to the model consists of the text $t_i$, user ID $u_i$ and/or the public perception score  $s_i$ for each turn in the conversation $i \in [1, 2, ..., N-1]$, as described below:

    \noindent \textbf{Textual input} --- the input consists of an encoding of the turn text $t_i$ using BERT embeddings extracted after fine tuning as described in \citet{dynamic}, resulting in the sequential encoding of the text as vector $t^{'}_{i}$ .
    
    \noindent \textbf{User ID input} --- the input consists of an encoding of the user ID as a randomly initialized vector $u_i$, where each user has a unique vector; we use  BiLSTM sequential encoding to obtain the utterance user ID vectors $u^{'}_{i}$. We use unique randomly initialized vectors to avoid privacy issues that may arise using actual IDs.
    
  \noindent \textbf{Public perception input} --- the input consists of a popularity score where the up-votes on a turn is subtracted by the down-votes on the same. To obtain the score vector $s_i$ we use equal depth binning to capture three levels of popularity for positive scores and three levels of unpopularity for negative scores. We use  a BiLSTM  sequential encoder on $s_i$  resulting in utterance public perception vectors $s^{'}_{i}$.

\subsection{Graph Construction} 

For a given conversation, the output of the sequential encoder for each one of these input types $t^{'}_{i}$,  $u^{'}_{i}$ and  $s^{'}_{i}$ is used to initialize the vertices in the homogeneous graphs shown in Figure \ref{GCN}. The vertices in the graphs represent the turns in the conversation. Each graph $G_{x} = (V, E, R, W)$, for each type of input $x\in\{t,u,s\}$, is constructed with vertices $v_i  \in V$,  $r_{ij} \in E$ is the labeled edges
 between $v_i$ and $v_j$, the edge labels (relations) $\in R$  and $\alpha_{ij}$ is the weight of the labeled edge $r_{ij}$, with $0 \leq \alpha_{ij}\leq 1$, where $ \alpha_{ij} \in  W$, $i\in [1, 2, ..., N-1]$ and $j \in$ the set of all neighboring vertices to $v_i$.

For each conversation we construct three types of graphs; a text-based $G_t$, a user-based  $G_u$ and a public perception score-based $G_s$. In $G_t$, each conversation turn is represented as a vertex $v_i \in V $ and is initialized with the textual sequentially encoded feature vector $t^{'}_{i}$, for all $i \in [1, 2, ..., N-1 ]$. In $G_u$ each vertex is initialized with a utterance user ID vector $u^{'}_{i}$, for all $i \in [1, 2, ..., N-1]$ provided by the sequential encoder. Similarly, in $G_s$ each vertex is initialized with a utterance public score vector $s^{'}_{i}$, for all $i \in [1, 2, ..., N-1 ]$ provided by the sequential encoder.

\begin{figure}[t]
\centering
   \includegraphics[width= 8cm]{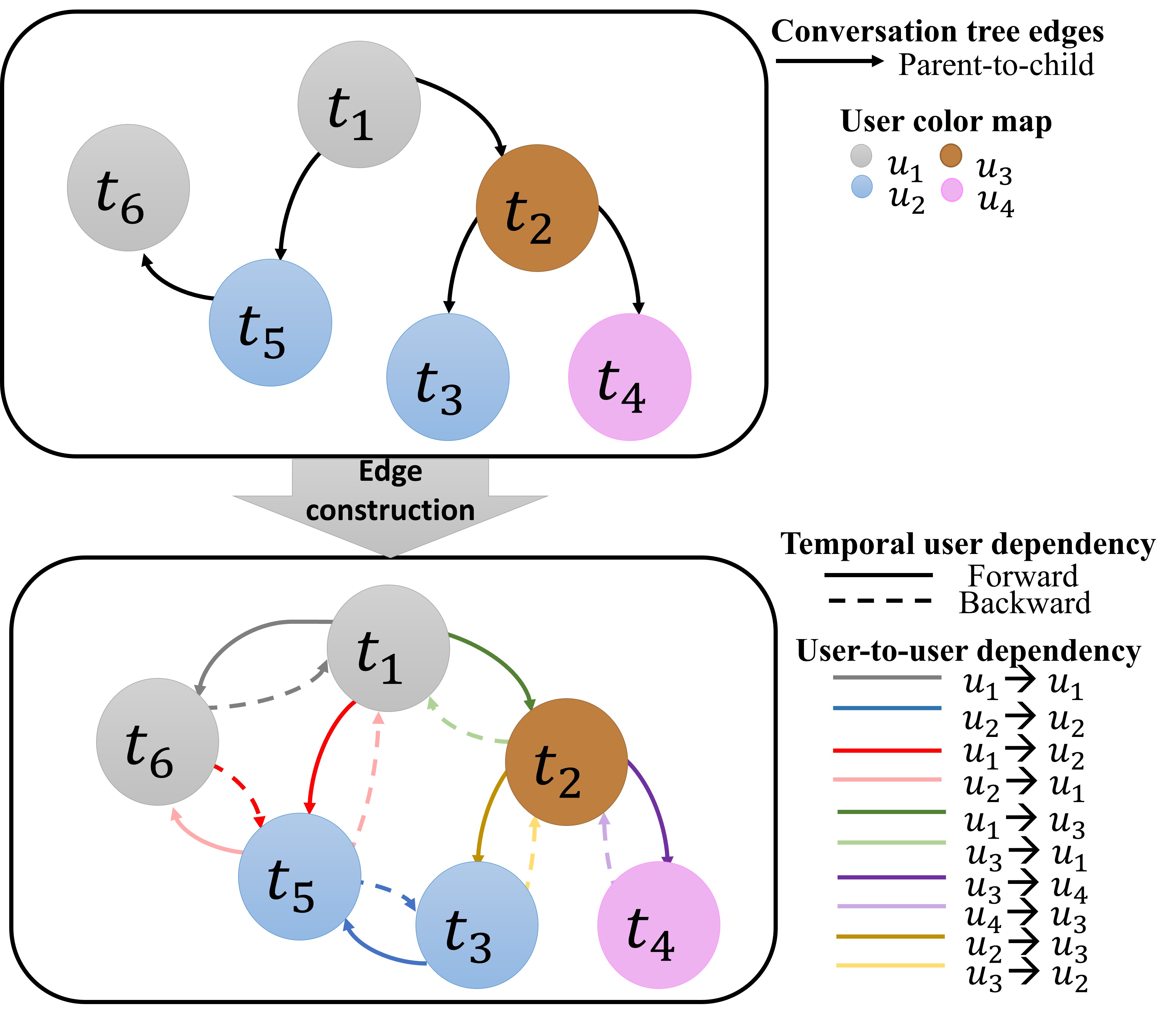}
   \caption{An example graph edge construction for the given conversation with user-to-user and temporal dependency.}
   \label{edge-construct}
   \vspace{-10pt}
\end{figure}

\smallskip\noindent\textbf{User to user relationship edge construction}. We establish the direct user to user relationship in a conversation through the edge construction of each graph.  This results in efficient graph modeling with less complexity compared to complete graphs.  Figure \ref{edge-construct} shows an example graph edge construction for a given conversation. In user-to-user relationship edge construction, each vertex $v_i$ representing a turn in the conversation has directed edges connecting it to its preceding (parent) and succeeding comments/turns (children). Same user comments (turns) are also connected through directed edges. The user-to-user relation $ \in R$ of an edge $r_{ij}$ is set based on the user-to-user dependency between user $u_{i}$ (of turn $v_i$) and user $u_{j}$ (of turn $v_j$). 

For example, in Figure \ref{edge-construct}, there are four users,  so the set of  edge labels $R$ has the relation types shown under user-to-user dependency. Furthermore, as the graph is directed, two vertices can have edges in both directions with different relations. We use this to represent the past (backward) and future (forward) temporal dependency between the vertices, shown as temporal user dependency.
The edge weights are set using a similarity based attention module. The attention function is computed such that, for each vertex, the incoming set of edges has a sum total weight of 1. The weights are calculated as, $\alpha_{ij} = softmax(v^T_{i} W_{e}[v_{j_{1}}, . . . , v_{j_{m}} ])$, for $j_k$, where $k= 1, 2, . . , m$, for the $m$ vertices connected to $v_i$, ensuring the vertex $v_i$ receives a total weight contribution of 1.

\subsection{Feature Transformation}
The sequentially encoded text features $t^{'}_{i}$, the user features  $u^{'}_{i}$, and the public perception features $s^{'}_{i}$ of the graph network are transformed from user dynamic independent into user dynamic dependent feature vectors using a two-step graph convolution process employed by \citet{DGCN}. In the first step, a new feature vector is computed for each vertex in all three graphs by aggregating local neighbourhood information:
\[ u^{''}_{i} = \sigma( \sum\limits_{r\in R} \sum\limits_{j \in N^r_i} \frac{\alpha_{ij}}{c_{i,r}} W_r\, u^{'}_{j}+\alpha_{ii}W_{0}\,u^{'}_{i}), \]

\[ t^{''}_{i} = \sigma( \sum\limits_{r\in R} \sum\limits_{j \in N^r_i} \frac{\alpha_{ij}}{c_{i,r}} W_r\, t^{'}_{j}+\alpha_{ii}W_{0}  \, t^{'}_{i}) \]

\[ s^{''}_{i} = \sigma( \sum\limits_{r\in R} \sum\limits_{j \in N^r_i} \frac{\alpha_{ij}}{c_{i,r}} W_r\, s^{'}_{j}+\alpha_{ii}W_{0}  \, s^{'}_{i}) \]
for $i = 1, 2, . . . , N-1,$ where, $\alpha_{ii}$ and $\alpha_{ij}$ are the edge weights, and $N^r_i$ is the neighbouring indices of vertex $i$ under relation $r \in R$. $c_{i,r}$ is a problem specific normalization constant automatically learned in a gradient based learning setup. $\Sigma$ is an activation function such as ReLU, and $W_{r}$ and $W_{0}$ are learnable parameters of the transformation. In the second step, another local neighbourhood based transformation is applied over the outputs of the first step, as follows:

\[ u^{''}_{i} = \sigma(  \sum\limits_{j \in N^r_i} W\, u^{''}_{j}+\alpha_{ii}W_{0}   \,u^{''}_{i}),\]

\[ t^{''}_{i} = \sigma( \sum\limits_{j \in N^r_i}  W\, t^{''}_{j}+\alpha_{ii}W_{0}  \, t^{''}_{i}) \]

\[ s^{''}_{i} = \sigma( \sum\limits_{j \in N^r_i}  W\, s^{''}_{j}+\alpha_{ii}W_{0}  \, s^{''}_{i}) \]
for $i = 1, 2, . . . , N-1$, where, $W$ and $W_{0}$ are transformation parameters, and $\sigma$ is the activation function. This two step transformation accumulates the normalized sum of the local neighbourhood.

\subsection{Forecasting Derailment }
To form the final turn representation, the sequential encoded vectors $t^{'}_{i}$, $u^{'}_{i}$ and $s^{'}_{i}$, and the user dynamic encoded vectors $t^{''}_{i}$, $u^{''}_{i}$ and $s^{''}_{i}$ are concatenated for each turn $i$ in a conversation to form:
$$g_{i} = [t^{'}_{i}, u^{'}_{i},s^{'}_{i},t^{''}_{i}, u^{''}_{i},s^{''}_{i}]$$ 
Then, each $g_i$, $i\in \{1,2....,N-1\}$ is concatenated to form a representation of the conversation $C$:
$$C^{'} = [g_1,g_{2}...,g_{N-1} ]$$
Finally, $C^{'}$ is fed to  a classifier with a linear layer, a full connected network and a sigmoid activation function, as described by \citet{DGCN}, to obtain the label $\hat{l}$ of each conversation $C$.

\subsection{Model variants}

To understand the effect of each type of input on forecasting derailment, we create variants of our model where the types of input are gradually included. The following is a more detailed description of the models used in this work.

    \smallskip\noindent\textbf{FGCN-T} --- this variant of the model constructs one graph using the output of the sequential encoder on the textual data $t^{'}_{i}$. It is created for both CGA and CMV, as both contain the textual data. FGCN-T uses only the textual layer shown in Figure \ref{GCN}. At classification it concatenates $t^{'}_{i}$ with the result of the GCN feature transformation $t^{''}_{i}$ during user dynamic encoding.
    
    \smallskip\noindent\textbf{FCGN-TU} --- this variant of the model constructs two graphs using the output of the sequential encoder, one for the textual output $t^{'}_{i}$ and the other for the user ID output $u^{'}_{i}$. It is created for both CGA and CMV, as both contain the textual and user ID data. FGCN-TU uses  the textual and user ID layer shown in Figure \ref{GCN}. At classification it concatenates $t^{'}_{i}$ and $u^{'}_{i}$ with the result of the GCN feature transformation $t^{''}_{i}$ and $u^{''}_{i}$ during user dynamic encoding.
    
   \smallskip \noindent\textbf{FGCN-TS} --- this variant of the model constructs two graphs using the output of the sequential encoder, one for the textual output $t^{'}_{i}$ and the other for the public perception score output $s^{'}_{i}$. It is created for CMV, as only CMV contains the public perception data. FGCN-TS uses  the textual and public score layer shown in Figure \ref{GCN}.  At classification it concatenates $t^{'}_{i}$ and $s^{'}_{i}$ with the result of the GCN feature transformation $t^{''}_{i}$ and $s^{''}_{i}$ during user dynamic encoding.
    
    \smallskip\noindent \textbf{FGCN-TSU} --- this variant of the model constructs three graphs using the output of the sequential encoder, one for the textual output $t^{'}_{i}$, the user ID output $u^{'}_{i}$ and the public perception score output $s^{'}_{i}$. It is created for CMV, as only CMV contains the public perception data. FGCN-TSU uses all layers shown in Figure \ref{GCN}. At classification it concatenates $t^{'}_{i}$, $u^{'}_{i}$ and $s^{'}_{i}$ with the result of the GCN feature transformation $s^{''}_{i}$ and $s^{''}_{i}$ during user dynamic encoding.

\begin{table}[!t]
\centering
\begin{tabular}{lcccrrr} 
\toprule
\textbf{Dataset} & \multicolumn{3}{c}{\textbf{Input}} & \textbf{Train} & \textbf{Val} & \textbf{Test}\\
&$t$&$u$&$s$&&&\\
\midrule
CGA &\checkmark  &\checkmark & \ding{53} & 2508& 840& 840\\
CMV & \checkmark &\checkmark &\checkmark & 4106 & 1368 & 1368\\

\bottomrule
\end{tabular}
\caption{Statistics of the datasets. $t$ denotes text input,  $u$ denotes user ID input and $s$ denotes public perception score input. All splits are balanced between the two classes.}\label{datasets}
\end{table}

\begin{table*}[!t]
\centering
\begin{tabular}{>{\centering\arraybackslash}p{1.5cm}|p{2.2cm}|>{\centering\arraybackslash}p{1cm}>{\centering\arraybackslash}p{1cm}>{\centering\arraybackslash}p{1cm}>{\centering\arraybackslash}p{1cm}|>{\centering\arraybackslash}p{1cm}>{\centering\arraybackslash}p{1cm}>{\centering\arraybackslash}p{1cm}>{\centering\arraybackslash}p{1cm}} 
\toprule
&&\multicolumn{4}{c|}{\textsc{\textbf{CGA}}} &\multicolumn{4}{c}{\textsc{\textbf{CMV}}}\\ 

\textsc{\textbf{Traning  }}&\centering\textsc{\textbf{Model}}&Acc& P& R &F1 &Acc &P &R& F1\\
\bottomrule

\multirow{6}{*}{\rotatebox{90}{\textsc{\textbf{Static}}}}&CRAFT &64.4& 62.7& 71.7 &66.9&  60.5& 57.5 &81.3 &67.3\\
&BERT·SC &64.7& 61.5& 79.4 &69.3&  62.0& 58.6& 82.8& 68.5\\

&FGCN-T & 66.4 & 63.0& 79.5&70.3 & 62.9 & 59.2& 83.0&  69.1  \\

&FGCN-TU &66.9 & 63.3&80.2  & \textbf{70.8} & 63.2 & 59.5& 83.0& 69.3\\ 
&FGCN-TS &-& - & - & - & 64.2&60.3 & 83.2& 69.9\\
&FGCN-TSU &-& - & - & - & 64.7& 60.7& 83.3& \textbf{70.2}\\

\bottomrule

\multirow{6}{*}{ \rotatebox{90}{\textsc{\textbf{Dynamic}}}} &BERT·SC+ &64.3& 61.2& 78.9& 68.8& 56.5& 56.0& 73.2& 61.7 \\

&FGCN-T+ & 65.7 &62.2 & 79.7&69.9 & 62.1 &58.5 &82.0&  68.3  \\

&FGCN-TU+ &65.9 &62.4 & 80.2& \textbf{70.2} & 62.7 & 58.8& 82.7& 68.8\\ 
&FGCN-TS+ &-& - & - & - &62.9 &59.2 &82.9 & 69.1\\
&FGCN-TSU+ &-& - & - & - & 63.5&59.7 & 83.1& \textbf{69.5}\\
\bottomrule
\end{tabular}
\caption{ Experimental results for forecasting conversation derailment. Best F1-score are in bold.}\label{f1-score}
\end{table*}

\section{Experimental Setup}
 \label{sec:experiments}

\subsection{Datasets}
We use two datasets for the task of forecasting derailment in conversations. Some statistics of the datasets are summarized in Table~\ref{datasets}.

\medskip
\noindent\textbf{Conversations Gone Awry (CGA}) dataset  \cite{CGA} was extracted from Wikipedia Talk Page conversations. The conversations were sampled from WikiConv \cite{wiki} based on an automatic measure of toxicity that ranges from 0 (not toxic) to 1 (is toxic). A conversation is extracted as a sample of derailment if the $N$th comment in it has a toxicity score higher than 0.6 and all the preceding comments have a score lower than 0.4. Conversations having all comments with a toxicity score below 0.4 are extracted as  samples of non-derailment. This set of conversations is further filtered through manual annotation to determine whether after an initial civil exchange a personal attack occurs from one user towards another. The conversations include the turn with the 
personal attack. This means all $N-1$ turns in a conversation are civil and the $N$'th one is either civil or contains a personal attack. The dataset also contains additional information about each comment in the conversation such as the user posting the comment and the ID  of the parent comment that this comment was a reply to. 

\medskip
\noindent\textbf{Reddit ChangeMyView (CMV)} dataset \cite{cmv} was extracted from Reddit conversations held under the ChangeMyView subreddit. Conversations  were identified  as derailed if there was a deletion of a turn by the platform moderators. This could have been done under Reddit’s Rule: ``Don’t be rude or hostile to other users.'' Unlike CGA, there is no control to ensure that all the preceding comments to the last one would be civil, resulting in some noise in the data. The dataset also contains additional information about each comment in the conversation such as the user posting the comment, the ID  of the parent comment that this comment was a reply to, and a votes score (i.e., the number of up-votes minus the number of down-votes). 


\subsection{Evaluation Metrics}

Following prior work, we report the performance of the models in terms of accuracy (Acc), precision (P), recall (R), and F1-score. We also report the forecast horizon H introduced by \citet{dynamic}, which is the mean of the turns in which the first detection of derailment occurred for the set of conversations that derail.

\subsection{Baselines}
Our FGCN model and its variants are evaluated against the state-of-the-art methods below.

\smallskip\noindent \textbf{CRAFT} \cite{cmv} is a model with a hierarchical recurrent neural network architecture, which integrates a generative dialog model that learns to represent conversational dynamics in an unsupervised fashion, and a supervised component that fine-tunes this representation to forecast future events.
This model is trained statically. Static training entails that all $N-1$ turns $\{t_{1}, ..., t_{N-1}\}$ of a conversation of $N$ turns are used as one input instance. 

\smallskip\noindent \textbf{BERT·SC} \cite{dynamic} is a model consisting of the BERT checkpoint with a sequence classification (SC) head, trained statically, i.e. in the same manner as CRAFT. 

\smallskip\noindent \textbf{BERT·SC+} \cite{dynamic} similar to BERT·SC consists of the BERT checkpoint with a sequence classification (SC) head, but is instead trained {\em dynamically}. Dynamic training entails that a conversation of $N$ turns $\{t_{1}, ..., t_{N-1}\}$ with label $l$ is mapped to multiple training samples, each representing a different phase of the conversation unfolding, but all labeled with the same label $l$. So a conversation of $N$ turns is converted into $N-1$
training instances samples $\{t_{1}\}, \{t_{1}, t_{2}\}, ... , \{t_{1}, ..., t_{N-1}\}$ instead of just the one $\{t_{1}, ..., t_{N-1}\}$.

\subsection{Implementation}
We use two training paradigms, static and dynamic: In \textbf{static training}, for each conversation we use one training instance with all turns $\{\{t_{1}, ..., t_{N-1}\}$, $\{u_1, ..., u_{N-1}\}$ and/or $\{s_1, ..., s_{N-1}\}\}$ as input. In \textbf{dynamic training}, we use multiple instances of each conversation, by varying the last turn used in each training instance. So, we use $\{t_{1}$, $u_{1}$ and/or $s_{1}\}$  as an instance,  $\{\{t_1, t_2\}$, $\{u_1, u_2\}$ , and/or $\{s_1, s_2\}\}$ as another instance,  and so on until the last instance  $\{\{t_{1}, ..., t_{N-1}\}$, $\{u_1, ..., u_{N-1}\}$ and/or $\{s_1, ..., s_{N-1}\}\}$. So we have $N-1$ instances of each conversation. We denote all dynamically trained models with an added ``+'' at the end of the model name.

At inference time, the model is tested dynamically, i.e., by using turn $\{t_{1}$, $u_{1}$ and/or $s_{1}\}$ as input, and making a prediction $\hat{l_{1}}$, then using turns $\{\{t_1, t_2\}$, $\{u_1, u_2\}$, and/or $\{s_1, s_2\}\}$, and making a prediction $\hat{l_{2}}$, and so on until $N-1$ predictions have been accumulated. The overall predicted label for a given conversation is then obtained as $\hat{l} = \max^{N-1}_{i=1} \hat{l_{i}}$.  

Our models used the same BERT implementation (i.e., \texttt{bert-base-uncased}) as in the baseline models \cite{dynamic}, for our textual sequential encoding, to ensure a comparable evaluation setting. However, it is worth mentioning that any pretrained language model can be used. The graph neural network component described in this work is implemented with settings similar to that reported  by \citet{DGCN}. The results are reported as an average over 10 different runs with random initialization, to account for variance in model performance.

\begin{figure*}[t!]
\centering
\includegraphics[width=0.96\textwidth]{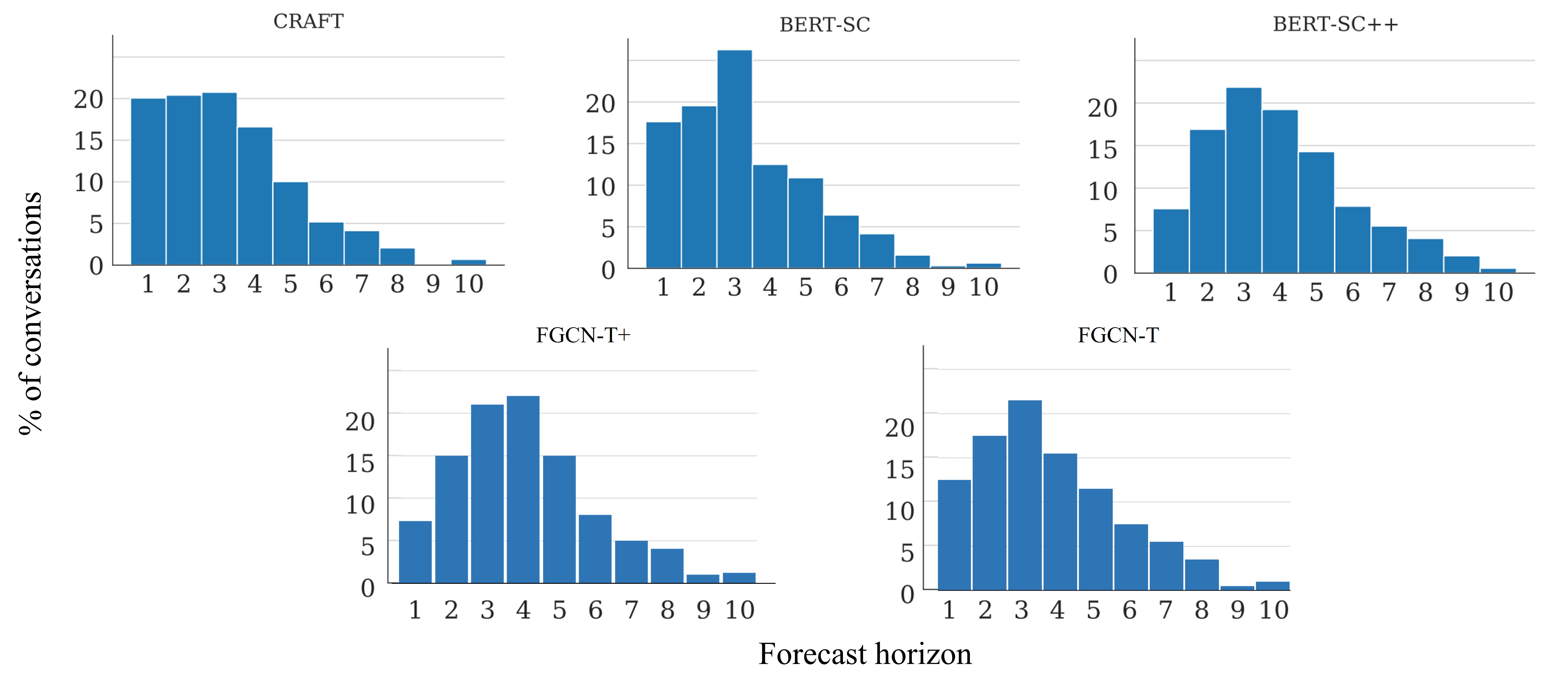}
 \caption{Forecast horizon on the CGA dataset with a model drawn at random from among the 10 available ones. A horizon of 1 means that an upcoming derailment was only predicted on the last turn before it occurred.}
\label{Hw}
\end{figure*}

\section{Results and Discussion} 
 \label{sec:results}

 In the  forecasting conversation derailment experiment we report the results of the static and dynamic training of our model and its variants and  compare  with baselines. In the analyzing mean forecast horizon experiment we show how early each model can forecast derailment. 
\subsection{Forecasting Conversation Derailment}

The results in Table \ref{f1-score} show that across both the datasets, FGCN-T, our text based graph neural network model, outperforms  baseline models when given similar text input. FGCN-TU, which explicitly uses user ID's in addition to text data, further improves the results for both the datasets. FGCN-TS, which uses text and public perception data in the CMV dataset, brings similar improvements. Furthermore, the results of the FGCN-TSU model, which uses text, user and public perception data, indicate that incorporating all three features when available can be beneficial. Note that CGA does not provide any public perception data and was excluded from this experiment.

Statically trained models outperform their corresponding dynamically trained models. However, our dynamically trained  models outperform  both statically and dynamically trained baselines. 

Taken together, these results indicate that modeling conversations using a graph neural network improves the models' forecasting F1-score. They also demonstrate that this modeling framework is flexible and allows for incorporating more types of data that may be beneficial. For instance, future work could investigate the potential benefit of integrating  explicit emotion or sentiment values \cite{babanejad2019leveraging, agrawal2016detecting} into derailment forecasting models.

\begin{table}[t]
\centering
\begin{tabular}{p{2.5cm}>{\centering\arraybackslash}p{1.9cm}>{\centering\arraybackslash}p{1.9cm}} 
\toprule
&\textbf{CGA}&\textbf{CMV}\\ 
\midrule
CRAFT & 2.36& 4.01\\
BERT·SC &2.60& 3.90\\
BERT·SC+ &\underline{2.85}&  \underline{4.06}\\
FGCN-T &2.73&  4.03\\
FGCN-T+ &\textbf{2.96}&  \textbf{4.12}\\
  \bottomrule
\end{tabular}
\caption{ Experimental results of mean forecast horizon (H).The best result is shown in bold whereas the second best result has been underlined. }\label{Horizon}
\vspace{-10pt}
\end{table}

\subsection{Analyzing Mean Forecast Horizon}

 How early can the model forecast the derailment? To answer this question we calculate the forecast horizon $H$, the mean of the turn in which the first detection of derailment occurred for the set of conversations that derail. A forecast horizon $H$  of 1 means that a derailment coming up on turn $N$ was first detected on turn $N-1$. A longer forecast horizon (i.e., a higher H) allows for earlier interventions and potentially allows moderators to delete the upcoming personal attack as soon as it appears on their platform to avoid any form of escalation. Models that are able to detect a potential intervention earlier have a clear advantage.

In Table \ref{Horizon} we report the results of the mean forecast horizon $H$. The results show that our model  FGCN-T+ with its dynamic training provides the earliest overall forecasting of derailment with a mean H  of 4.12 for CMV, and  2.95 for CGA. Followed by another  dynamically trained model BERT.SC+.   For the statically trained models (CRAFT, BERT.SC, and FGCN-T), FGCN-T has the best performance as it seems to be able to better model the dynamic relationships between the users of the turns with its graph model, obtaining a mean H of 4.03 for CMV and,  2.73 for CGA.  



We perform further analysis of the  forecast horizon results for CGA. Figure~\ref{Hw} illustrates the forecast horizon of the different models. In earlier work,  both CRAFT and BERT·SC, the models make a prediction with a forecast horizon $H>1$ turn at a high rate. Only 20\% of CRAFT’s forecasts and 17.5\% of BERT·SC’s forecasts came on the last turn before the derailment. Turning to BERT·SC+, we see that dynamic training has helped in shifting a lot of the density from $H <= 3$ towards $H > 3$. The last-minute forecasts for BERT·SC+ model come at a rate of only 7.5\%. FGCN-T and FGCN-T+ uses BERT.SC in its sequential textual component and combines it with a graph component. FGCN-T  was also able to shift density  from $H <= 3$ towards $H > 3$ even though it was trained statically, indicating that the result was due to the graph modeling.   The dynamiclly trained  FGCN-T+ outperforms all by  shifting even more density towards   $H > 3$ and a last minute forecast at a rate of only 7\%.


\section {Conclusion}
\label{sec:conclusions}
Unlike previous models which were based on simpler sequence models, FGCN is built on a graph convolutional neural network and is able to capture the dynamics of multi-party dialogue, including user relationships and public perception of conversation statements. FGCN performed significantly better than state-of-the-art models on two widely used benchmark datasets, CGA and CMV. Conversation derailment is a significant issue that frequently and severely impacts our online social interactions, whether in casual settings or more formal contexts such as online learning or remote work. The ability to accurately predict derailment has the potential to enhance the effectiveness of moderation and thus protect individuals who are vulnerable to emotional abuse or harm and improve the overall quality of online interactions.


\section*{Limitations}

Graph models require four or more utterances to form meaningful conversation connections and model their dynamics. In some cases, conversations that derail are not sufficiently long and may be best modeled by simpler sequential models. Any of these models will work best with asynchronous conversations where there is a time lag between the turns to allow for moderation after forecasting.

\section*{Ethics Statement}
In our paper, we focus on the problem of forecast-
ing conversation derailment. The practical employ-
ment of any such system on online platforms has
potential positive impact, but several things would
be important to first consider, including whether
forecasting is fair \cite{ethics1},
how to inform users about the forecasting (in ad-
vance, and when the forecasting affects users), and
finally what other action is taken when derailment
is forecast. Please refer to \cite{ethics2} for a
related overview of such considerations, in the context of abusive language detection.

\section*{Acknowledgments}
We are grateful to the anonymous reviewers for their insightful feedback. The financial support of NSERC, the NCRN, VISTA and the IDEaS network is gratefully acknowledged.

\bibliography{anthology,custom}
\bibliographystyle{acl_natbib}


\end{document}

%% file: introduction.tex
\section{Introduction}

The widespread availability of chat or messaging platforms, social media, forums and other online communities has led to an increase in the number of online conversations between individuals and groups. 
In contrast to offline or face-to-face communication,  online conversations require moderation to maintain the integrity of the platform and protect users' privacy and safety \cite{virtual}. Moderation can help to prevent harassment, trolling, hate speech, and other forms of abusive behavior \cite{thirty}. It can also help to prevent and address conversation derailment.

\begin{figure}[t]
     \centering
     \begin{subfigure}[b]{0.25\textwidth}
         \centering
         \includegraphics[width=\textwidth]{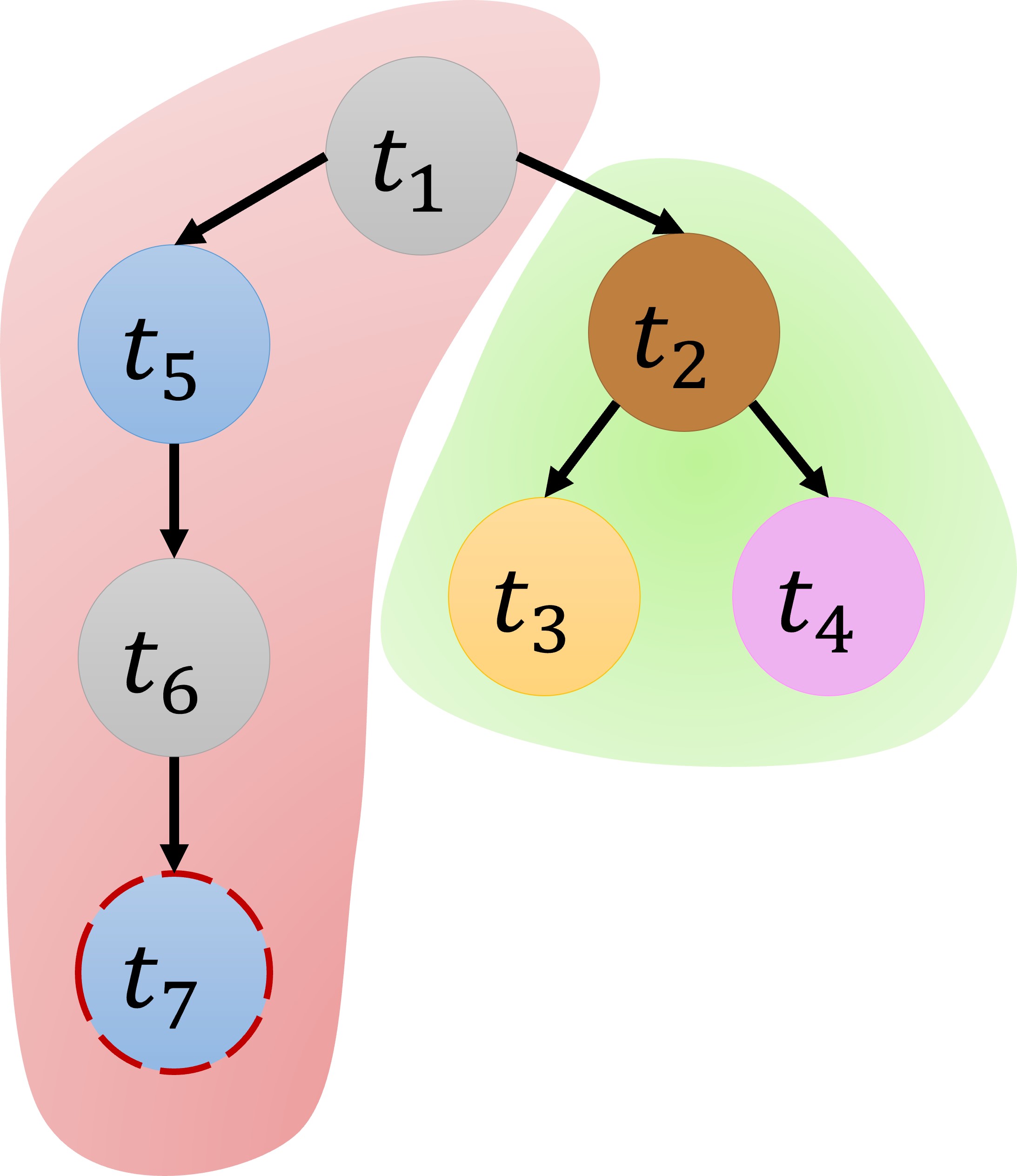} 
         \caption{Conversation $C_{CGA}$}
         \label{fig:example}
     \end{subfigure}
     \begin{subfigure}[b]{0.22\textwidth}
         \centering
         \includegraphics[width=\textwidth]{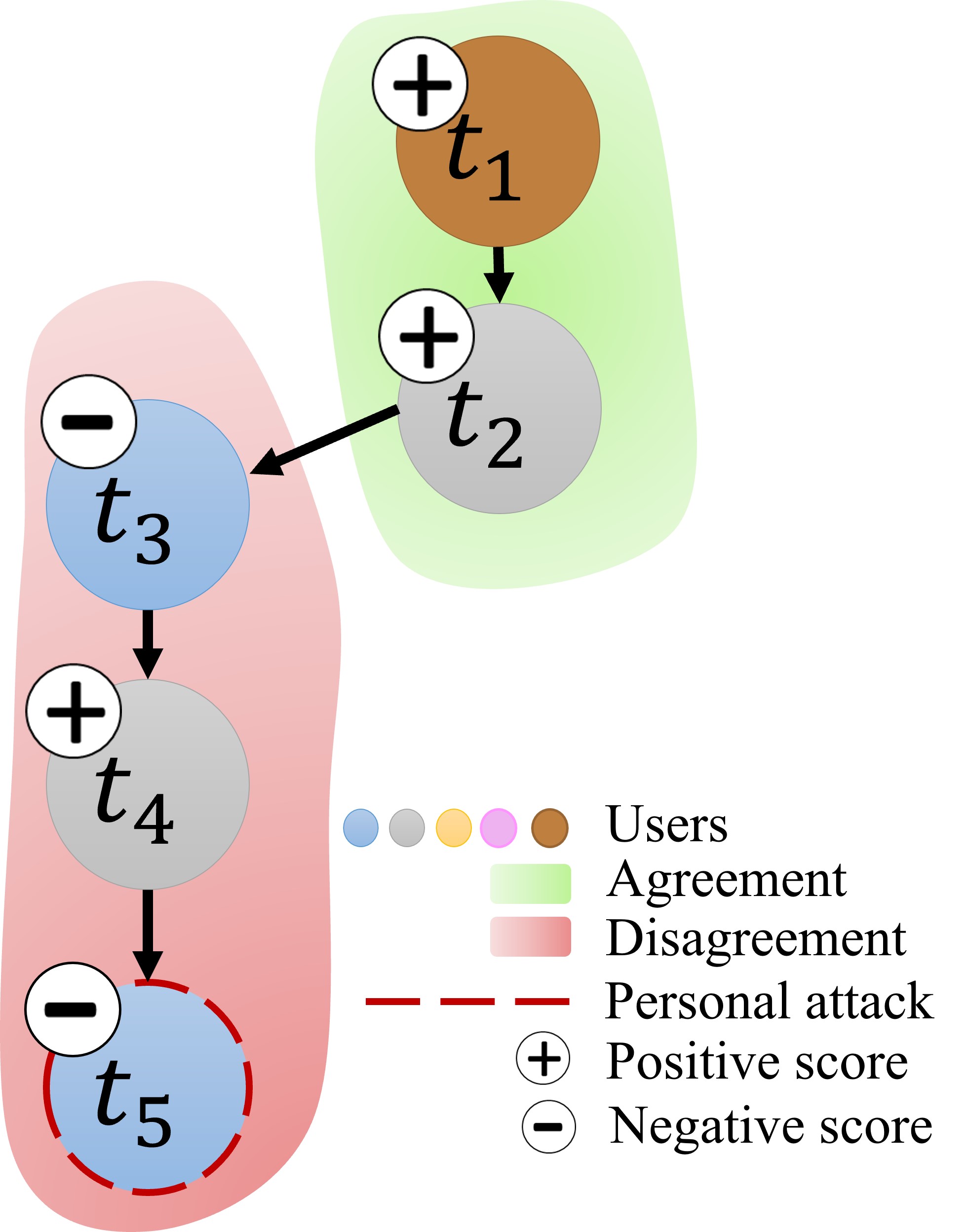}
         \caption{Conversation $C_{CMV}$}
         \label{fig:examplecmv}
     \end{subfigure}
\caption{Conversation derailment examples coming from two benchmark datasets, CGA and CMV; (a) illustrates graph dynamics in a conversation, and (b) illustrates public perception through votes in a conversation. Our FGCN model exploits these features to improve the accuracy of conversation derailment forecasting.}
\label{fig:cm}
\end{figure}

{\em Conversation derailment} refers to the process by which a conversation or discussion is redirected away from its original topic or purpose, typically as a result of inappropriate or off-topic comments or actions by one or more participants. In online conversations, derailment can be exacerbated by the lack of nonverbal cues and the perceived anonymity that can be provided by the internet. Conversation derailment can lead to confusion, frustration, and a lack of productivity or progress in the conversation.
Table \ref{tab:example} shows an example conversation taken from the popular CGA benchmark dataset \cite{CGA}. One can observe that there is offensive language used by one of the participants that leads the conversation to derail. The severity of the verbal attack may indicate a prior history between the two participants in previous conversations.

\begin{table}[t!]
\small
\centering
\begin{tabular}
{ccp{3.7cm}c}
\toprule
\textbf{Turn} & \textbf{User} & \textbf{Text} & \textbf{Label}\\  
\midrule
$N-3$  & $\mathcal{A}$ & {\em ``Proper use of an editor's history includes  fixing errors or violations of Wikipedia policy or correcting related problems on multiple articles." } &   \\
$N-2$ & $\mathcal{B}$ & {\em  ``It's very clear that you just go to my contributions list and look to see what biography articles I've worked on, then you go and look to see if you can find something wrong with them.  
" } & \\
$N-1$ & $\mathcal{A}$ & {\em  ``So, what is wrong with fixing things? At the top of my talk page, it says to keep it on your watchlist. "} & \\
$N$ & $\mathcal{B}$ & {\em  ``You cannot possibly be too stupid to understand the warning I'm giving you.  I'm not going to repeat it."}  & \textbf{\textbf{?}}\\
\bottomrule
\end{tabular}
\caption{A sample conversation from the Conversation Gone Awry (CGA) dataset showing a sequence of text utterances that end with a verbal abuse. Given the conversation context up to $N-1$ turns, the task is to predict whether turn $N$ will be a respectful or offensive statement prior to it being presented leading to derailment (\textbf{it is offensive}, in this case).}
\label{tab:example}
\end{table}

In this research, we examine the problem of {\em forecasting conversation derailment}. The ability to predict conversation derailment early  has multifold benefits: (i) it is {\em more timely}, as it allows for proactive moderation of conversations (before they cause any harm) due to early warning, (ii) it is {\em more scalable}, as it allows to automatically monitor large active online communities, a task that is otherwise time-consuming, (iii) it is {\em more cost-effective}, as it may provide enough aid to limit the number of human moderators needed, and (iv) it may identify upsetting content early and prevent human moderators from being exposed to it. 

Early efforts towards automatic moderation focused on detecting inappropriate comments once they have occurred. But the utility of such an approach is limited as participants have already been exposed to an abusive behavior and any potential harm has already been caused. The current state-of-the-art approach to predict conversation derailment relies on sequence models that treat dialogues as text streams \cite{chang2022thread, dynamic}. 
However, this approach has limitations, as it ignores the semantics and dynamicity of a multi-party dialogue involving individuals' intrinsic tendencies and the history of interactions with other individuals.

Based on these observations, we propose a graph-based model to capture multi-party multi-turn dialogue dynamics. In addition, we leverage information associated with conversation utterances that provide public perception on whether an utterance is perceived as positive or negative.

There exist two popular benchmark datasets typically employed for the problem of interest: Conversations Gone Awry (CGA) \cite{CGA} and Reddit ChangeMyView (CMV) \cite{cmv}. Both datasets contain multi-party conversations, including the text and an anonymous user ID of each utterance, along with a label annotation on whether the conversation will derail or not. CMV also includes a public vote on each of the utterances that provides the public perception of individuals towards it.

Figure \ref{fig:example} shows an example multi-party conversation from CGA that is not sequential in nature. Note as well that some participants are in either agreement or disagreement of a heated argument. Graph models are more accustomed to represent such dialogue dynamics. Figure \ref{fig:examplecmv} shows an example conversation from CMV that shows sample voting scores on each utterance in the conversation that could be related to derailment. 

Graph neural networks have been successfully used to model conversations for downstream classification tasks. For example, they have shown promise in forecasting the next emotion in a conversation  \cite{predict}, a problem similar to that of interest in this work. This motivated us to explore this line of research and make the following contributions:
\begin{itemize}
    \item We propose a novel model based on a graph convolutional neural network, {\em the Forecasting Graph Convolutional Network (FGCN)}, that captures dialogue user dynamics and public perception of conversation utterances.
    \item We perform an extensive empirical evaluation of FGCN that shows it outperforms the state-of-the-art models on the GCA and CMV benchmark datasets by 1.5\% and 1.7\%, respectively. 
\end{itemize}

The remainder of the paper is organized as follows. Section~\ref{sec:related} reviews  related work. The technical problem of interest is presented in Section~\ref{sec:problem}. Section~\ref{sec:models} presents the proposed models. Section~\ref{sec:experiments} presents the experimental setup, and Section~\ref{sec:results} presents the results and a discussion. We conclude in  Section~\ref{sec:conclusions}.

%% file: acl_latex.bbl
\begin{thebibliography}{29}
\expandafter\ifx\csname natexlab\endcsname\relax\def\natexlab#1{#1}\fi

\bibitem[{Agrawal and An(2014)}]{agrawal2014kea}
Ameeta Agrawal and Aijun An. 2014.
\newblock Kea: Sentiment analysis of phrases within short texts.
\newblock In \emph{Proceedings of the 8th International Workshop on Semantic
  Evaluation (SemEval 2014)}.

\bibitem[{Agrawal et~al.(2016)Agrawal, Sahdev, Davoudi, Khonsari, An, and
  McGrath}]{agrawal2016detecting}
Ameeta Agrawal, Raghavender Sahdev, Heidar Davoudi, Forouq Khonsari, Aijun An,
  and Susan McGrath. 2016.
\newblock Detecting the magnitude of events from news articles.
\newblock In \emph{2016 IEEE/WIC/ACM International Conference on Web
  Intelligence (WI)}. IEEE.

\bibitem[{Babanejad et~al.(2019)Babanejad, Agrawal, Davoudi, An, and
  Papagelis}]{babanejad2019leveraging}
Nastaran Babanejad, Ameeta Agrawal, Heidar Davoudi, Aijun An, and Manos
  Papagelis. 2019.
\newblock Leveraging emotion features in news recommendations.
\newblock In \emph{Proceedings of the 7th International Workshop on News
  Recommendation and Analytics (INRA)}.

\bibitem[{Chang and Danescu-Niculescu-Mizil(2019)}]{cmv}
Jonathan~P. Chang and Cristian Danescu-Niculescu-Mizil. 2019.
\newblock \href {https://doi.org/10.18653/v1/D19-1481} {Trouble on the horizon:
  Forecasting the derailment of online conversations as they develop}.
\newblock In \emph{Proceedings of the 2019 Conference on Empirical Methods in
  Natural Language Processing and the 9th International Joint Conference on
  Natural Language Processing (EMNLP-IJCNLP)}, pages 4743--4754, Hong Kong,
  China. Association for Computational Linguistics.

\bibitem[{Chang et~al.(2022)Chang, Schluger, and
  Danescu-Niculescu-Mizil}]{chang2022thread}
Jonathan~P Chang, Charlotte Schluger, and Cristian Danescu-Niculescu-Mizil.
  2022.
\newblock Thread with caution: Proactively helping users assess and deescalate
  tension in their online discussions.
\newblock \emph{Proceedings of the ACM on Human-Computer Interaction},
  6(CSCW2):1--37.

\bibitem[{Davidson et~al.(2017)Davidson, Warmsley, Macy, and Weber}]{hate}
Thomas Davidson, Dana Warmsley, Michael~W. Macy, and Ingmar Weber. 2017.
\newblock \href {http://arxiv.org/abs/1703.04009} {Automated hate speech
  detection and the problem of offensive language}.
\newblock \emph{CoRR}, abs/1703.04009.

\bibitem[{De~Kock and Vlachos(2021)}]{de2021beg}
Christine De~Kock and Andreas Vlachos. 2021.
\newblock I beg to differ: A study of constructive disagreement in online
  conversations.
\newblock \emph{arXiv preprint arXiv:2101.10917}.

\bibitem[{Devlin et~al.(2018)Devlin, Chang, Lee, and Toutanova}]{bert}
Jacob Devlin, Ming{-}Wei Chang, Kenton Lee, and Kristina Toutanova. 2018.
\newblock \href {http://arxiv.org/abs/1810.04805} {{BERT:} pre-training of deep
  bidirectional transformers for language understanding}.
\newblock \emph{CoRR}, abs/1810.04805.

\bibitem[{Ghosal et~al.(2019)Ghosal, Majumder, Poria, Chhaya, and
  Gelbukh}]{DGCN}
Deepanway Ghosal, Navonil Majumder, Soujanya Poria, Niyati Chhaya, and
  Alexander~F. Gelbukh. 2019.
\newblock \href {http://arxiv.org/abs/1908.11540} {Dialoguegcn: {A} graph
  convolutional neural network for emotion recognition in conversation}.
\newblock \emph{CoRR}, abs/1908.11540.

\bibitem[{Hua et~al.(2018)Hua, Danescu-Niculescu-Mizil, Taraborelli, Thain,
  Sorensen, and Dixon}]{wiki}
Yiqing Hua, Cristian Danescu-Niculescu-Mizil, Dario Taraborelli, Nithum Thain,
  Jeffery Sorensen, and Lucas Dixon. 2018.
\newblock \href {https://doi.org/10.18653/v1/D18-1305} {{W}iki{C}onv: A corpus
  of the complete conversational history of a large online collaborative
  community}.
\newblock In \emph{Proceedings of the 2018 Conference on Empirical Methods in
  Natural Language Processing}, pages 2818--2823, Brussels, Belgium.
  Association for Computational Linguistics.

\bibitem[{Janiszewski et~al.(2021)Janiszewski, Lango, and Stefanowski}]{time}
Piotr Janiszewski, Mateusz Lango, and Jerzy Stefanowski. 2021.
\newblock \href {https://doi.org/10.1007/978-3-030-86517-7_22} {Time aspect in
  making an actionable prediction of a conversation breakdown}.
\newblock page 351–364, Berlin, Heidelberg. Springer-Verlag.

\bibitem[{Kementchedjhieva and S{\o}gaard(2021)}]{dynamic}
Yova Kementchedjhieva and Anders S{\o}gaard. 2021.
\newblock \href {https://doi.org/10.18653/v1/2021.emnlp-main.624} {Dynamic
  forecasting of conversation derailment}.
\newblock In \emph{Proceedings of the 2021 Conference on Empirical Methods in
  Natural Language Processing}, pages 7915--7919, Online and Punta Cana,
  Dominican Republic. Association for Computational Linguistics.

\bibitem[{Kilvington(2021)}]{virtual}
Daniel Kilvington. 2021.
\newblock The virtual stages of hate: Using goffman’s work to conceptualise
  the motivations for online hate.
\newblock \emph{Media, Culture \& Society}, 43(2).

\bibitem[{Kiritchenko et~al.(2020)Kiritchenko, Nejadgholi, and
  Fraser}]{ethics2}
Svetlana Kiritchenko, Isar Nejadgholi, and Kathleen~C. Fraser. 2020.
\newblock \href {http://arxiv.org/abs/2012.12305} {Confronting abusive language
  online: {A} survey from the ethical and human rights perspective}.
\newblock \emph{CoRR}, abs/2012.12305.

\bibitem[{Li et~al.(2021)Li, Zhu, Li, Wang, Li, Liao, and Zheng}]{predict}
Dayu Li, Xiaodan Zhu, Yang Li, Suge Wang, Deyu Li, Jian Liao, and Jianxing
  Zheng. 2021.
\newblock Emotion inference in multi-turn conversations with addressee-aware
  module and ensemble strategy.
\newblock In \emph{Proceedings of the 2021 Conference on Empirical Methods in
  Natural Language Processing}, pages 3935--3941.

\bibitem[{Liang et~al.(2022)Liang, Meng, Zhang, Chen, Xu, and Zhou}]{HCNN}
Yunlong Liang, Fandong Meng, Ying Zhang, Yufeng Chen, Jinan Xu, and Jie Zhou.
  2022.
\newblock \href {https://doi.org/https://doi.org/10.1016/j.artint.2022.103714}
  {Emotional conversation generation with heterogeneous graph neural network}.
\newblock \emph{Artificial Intelligence}, 308:103714.

\bibitem[{Lubis et~al.(2019)Lubis, Sakti, Yoshino, and Nakamura}]{PositiveEE}
Nurul Lubis, Sakriani Sakti, Koichiro Yoshino, and Satoshi Nakamura. 2019.
\newblock Positive emotion elicitation in chat-based dialogue systems.
\newblock \emph{IEEE/ACM Transactions on Audio, Speech, and Language
  Processing}, 27:866--877.

\bibitem[{Salehabadi et~al.(2022)Salehabadi, Groggel, Singhal, Roy, and
  Nilizadeh}]{salehabadi2022user}
Nazanin Salehabadi, Anne Groggel, Mohit Singhal, Sayak~Saha Roy, and Shirin
  Nilizadeh. 2022.
\newblock User engagement and the toxicity of tweets.
\newblock \emph{arXiv preprint arXiv:2211.03856}.

\bibitem[{Saveski et~al.(2021)Saveski, Roy, and Roy}]{saveski2021structure}
Martin Saveski, Brandon Roy, and Deb Roy. 2021.
\newblock The structure of toxic conversations on twitter.
\newblock In \emph{Proceedings of the Web Conference}.

\bibitem[{Schluger et~al.(2022)Schluger, Chang, Danescu-Niculescu-Mizil, and
  Levy}]{schluger2022proactive}
Charlotte Schluger, Jonathan~P Chang, Cristian Danescu-Niculescu-Mizil, and
  Karen Levy. 2022.
\newblock Proactive moderation of online discussions: Existing practices and
  the potential for algorithmic support.
\newblock \emph{Proceedings of the ACM on Human-Computer Interaction},
  6(CSCW2):1--27.

\bibitem[{Sun et~al.(2021)Sun, Yu, and Fu}]{DisGNN}
Yang Sun, Nan Yu, and Guohong Fu. 2021.
\newblock \href {https://doi.org/10.18653/v1/2021.findings-emnlp.252} {A
  discourse-aware graph neural network for emotion recognition in multi-party
  conversation}.
\newblock pages 2949--2958.

\bibitem[{Tontodimamma et~al.(2021)Tontodimamma, Nissi, Sarra, and
  Fontanella}]{thirty}
Alice Tontodimamma, Eugenia Nissi, Annalina Sarra, and Lara Fontanella. 2021.
\newblock Thirty years of research into hate speech: topics of interest and
  their evolution.
\newblock \emph{Scientometrics}, 126(1):157--179.

\bibitem[{Wang and Cardie(2016)}]{sent}
Lu~Wang and Claire Cardie. 2016.
\newblock \href {http://arxiv.org/abs/1606.05704} {A piece of my mind: {A}
  sentiment analysis approach for online dispute detection}.
\newblock \emph{CoRR}, abs/1606.05704.

\bibitem[{Wiegand et~al.(2019)Wiegand, Ruppenhofer, and Kleinbauer}]{abuse}
Michael Wiegand, Josef Ruppenhofer, and Thomas Kleinbauer. 2019.
\newblock \href {https://doi.org/10.18653/v1/N19-1060} {{D}etection of
  {A}busive {L}anguage: the {P}roblem of {B}iased {D}atasets}.
\newblock In \emph{Proceedings of the 2019 Conference of the North {A}merican
  Chapter of the Association for Computational Linguistics: Human Language
  Technologies, Volume 1 (Long and Short Papers)}, pages 602--608, Minneapolis,
  Minnesota. Association for Computational Linguistics.

\bibitem[{Williamson and Menon(2019)}]{ethics1}
Robert Williamson and Aditya Menon. 2019.
\newblock \href {https://proceedings.mlr.press/v97/williamson19a.html}
  {Fairness risk measures}.
\newblock In \emph{Proceedings of the 36th International Conference on Machine
  Learning}, volume~97 of \emph{Proceedings of Machine Learning Research},
  pages 6786--6797. PMLR.

\bibitem[{Xie and Agrawal(2023)}]{xie23}
Justin Xie and Ameeta Agrawal. 2023.
\newblock Emotion and sentiment guided paraphrasing.
\newblock In \emph{Proceedings of the 13th Workshop on Computational Approaches
  to Subjectivity, Sentiment and Social Media Analysis}.

\bibitem[{Zhang et~al.(2018)Zhang, Chang, Danescu-Niculescu-Mizil, Dixon,
  Taraborelli, Thain, and Taraborelli}]{CGA}
Justine Zhang, Jonathan~P. Chang, Cristian Danescu-Niculescu-Mizil, Lucas
  Dixon, Dario Taraborelli, Nithum Thain, and Dario Taraborelli. 2018.
\newblock Conversations gone awry: Detecting warning signs of conversational
  failure.
\newblock In \emph{Proceedings of ACL}.

\bibitem[{Zhong et~al.(2019)Zhong, Wang, and Miao}]{affect-rich}
Peixiang Zhong, Di~Wang, and Chunyan Miao. 2019.
\newblock \href {https://doi.org/10.1609/aaai.v33i01.33017492} {An affect-rich
  neural conversational model with biased attention and weighted cross-entropy
  loss}.

\bibitem[{Ziems et~al.(2022)Ziems, Li, Zhang, and Yang}]{ziems2022inducing}
Caleb Ziems, Minzhi Li, Anthony Zhang, and Diyi Yang. 2022.
\newblock Inducing positive perspectives with text reframing.
\newblock \emph{arXiv preprint arXiv:2204.02952}.

\end{thebibliography}
